\documentclass[10pt,twocolumn,letterpaper]{article}

\usepackage[pagenumbers]{cvpr} 

\usepackage{graphicx}
\usepackage{amsmath}
\usepackage{amssymb}
\usepackage{booktabs}
\usepackage{multirow}

\usepackage[pagebackref,breaklinks,colorlinks]{hyperref}

\usepackage[capitalize]{cleveref}
\crefname{section}{Sec.}{Secs.}
\Crefname{section}{Section}{Sections}
\Crefname{table}{Table}{Tables}
\crefname{table}{Tab.}{Tabs.}

\def\cvprPaperID{*****} 
\def\confName{CVPR}
\def\confYear{2022}

\begin{document}

\title{TENET: Transformer Encoding Network for Effective Temporal Flow on Motion Prediction}
\author{Yuting Wang$^1$\footnotemark[1]\; \;Hangning Zhou$^1$\footnotemark[1]\; \;Zhigang Zhang$^1$\footnotemark[1]\; \;Chen Feng$^2$\; \;Huadong Lin$^3$\; \\ \;Chaofei Gao$^3$\; \;Yizhi Tang$^4$\; \;Zhenting Zhao$^{1,5}$\; \;Shiyu Zhang$^6$\; \;Jie Guo$^3$\; \\ \;Xuefeng Wang$^1$\; \;Ziyao Xu$^1$\; \;Chi Zhang$^1$\\
$^1$Megvii Inc. $^2$The Hong Kong University of Science and Technology $^3$Beihang University \\$^4$University of Michigan, Ann Arbor $^5$Harbin Institute of Technology $^6$Tsinghua University\\ 
\tt\small $^1$\{wangyuting, zhouhangning, zhangzhigang\}@megvii.com $^2$\{cfengag\}@connect.ust.hk \\ \tt\small $^4$\{linhuadong, 18376047, SGEguojie\}@buaa.edu.cn $^5$\{tangyz\}@umich.edu \\ \tt\small  $^3$\{zhang-sy21\}@mails.tsinghua.edu.cn $^1$\{zhaozhenting, wangxuefeng, xuziyao, zhangchi\}@megvii.com
}

\maketitle
\renewcommand{\thefootnote}{\fnsymbol{footnote}} 
\footnotetext[1]{The first three authors contribute equally to this work.}

\begin{abstract}
   This technical report presents an effective method for motion prediction in autonomous driving. We develop a Transformer-based method for input encoding and trajectory prediction. Besides, we propose the Temporal Flow Header to enhance the trajectory encoding. In the end, an efficient K-means ensemble method is used. Using our Transformer network and ensemble method, we win the \textbf{first} place of Argoverse 2 Motion Forecasting Challenge with the \textbf{state-of-the-art} brier-minFDE score of \textbf{1.90}.
\end{abstract}

\section{Introduction}
\label{sec:intro}
This technical report aims to share details of our method. Our network is developed upon SceneTransformer \cite{ngiam2021scene}, a Transformer-based \cite{xiong2020layer} model. For agent trajectories, time-wise and agent-wise self-attention is used to encode time sequence and spatial interaction information. Besides, cross-attention is used to share map information with agent trajectories. Finally, we predict $K$ trajectories and their corresponding scores from $K$ learnable tokens. These learnable tokens get history trajectories and map information from mixed features of agent and map through cross-attention layer.

In general, we summarize the contribution of our proposed algorithm as follow: 

\begin{itemize}
    \item We propose \textbf{Temporal Flow Header}
to enhance the flow of temporal information in the whole network.
    \item We propose a K-means method for the ensemble stage, and
achieve \textbf{\textit{state-of-the-art}} performance.
    \item For Transformer model, we propose an efficient strategy to reduce the input size in the training stage and increase the size in the testing stage to achieve faster training of the model while obtaining high accuracy.
\end{itemize}

\begin{figure*}[t]
  \centering
  \includegraphics[width=1.0\textwidth]{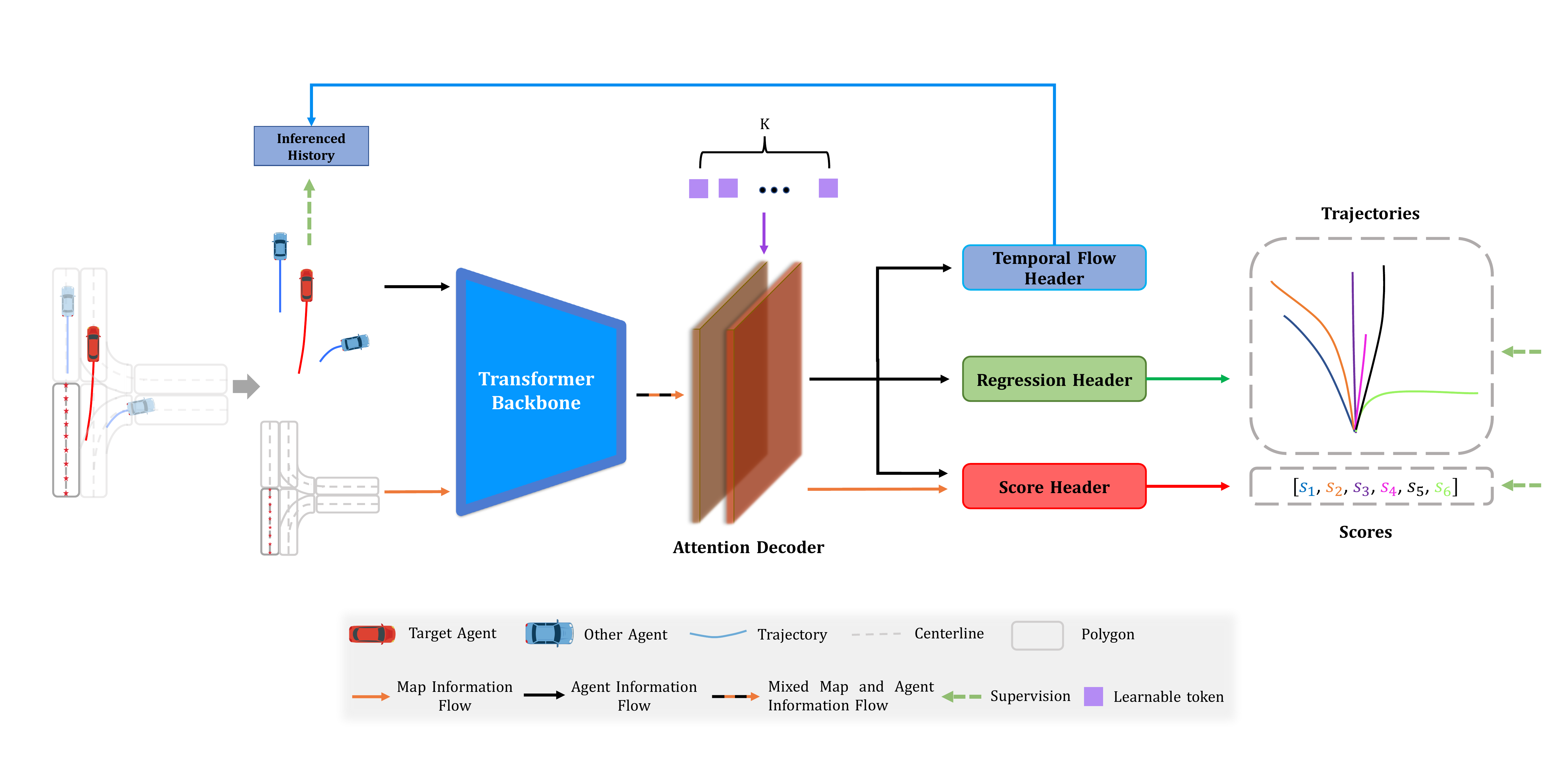}
  \setlength{\abovecaptionskip}{5mm}
  \caption{\textbf{Overall architecture of the proposed model TENET.} }
  \label{fig:fig1}
\end{figure*}

\section{Methodology}
\label{sec:formatting}

The overview of TENET is given in Fig. \ref{fig:fig1}. The proposed motion prediction model consists of three modules: (1) Transformer-based \cite{xiong2020layer} encoder that extracts spatially and temporally the feature of agent and map. (2) Attention decoder that utilizes learnable token \cite{carion2020end} to query effective information from mixed features of agent and map after encoder. (3) Three output headers that are for regression, score, and enhancing temporal information interaction.

\subsection{Model}

\noindent\textbf{Encoder and Decoder} We adopt efficient self-attention and cross-attention from SceneTransformer \cite{ngiam2021scene} to implement intra and inter information interaction of agent and map.
\begin{center}
    $SelfAtt_{i}(x) \gets Att_{i}(x, x, x),$ \\
    $CrossAtt_{i, j}(x1, x2) \gets Att_{i, j}(x1, x2, x2),$
\end{center}
where $SelfAtt_{i}(x)$ performs attention over axis $i$ of the tensor $x$, mixing information along axis $i$ while keeping information along other axes independent. It is the same for $CrossAtt_{i,j}(x1, x2)$. In decoder, we generate $K$ learnable tokens as queries to learn prediction trajectory feature from mixed features of agent and map after encoder through cross-attention layer,
\begin{center}
    $x_{pt} = SelfAtt_{K}(CrossAtt_{K,M}(x_{k}, x_{m})),$
\end{center}
where $x_{m}$ is the mixed feature of agent and map, $x_{k}$ is learnable token, and $x_{pt}$ is prediction trajectory feature.

\noindent\textbf{Regression Header} We take a 2-layer MLP to decode the trajectory information from $x_{pt}$ in all timestamps (including history and future), including positions, orientation, and instant velocity in BEV coordinates. To eliminate the effect of long-range prediction error accumulation on a predicted trajectory,  we use LSTM \cite{hochreiter1997long} to extract time-sequential information on target agent feature, and add it to post-MLP trajectory information. The final forecasted regression result is defined as
\begin{center}
    $Traj_{pred} = MLP(x_{pt}) + LSTM(x_{pt}).$
\end{center}

\noindent\textbf{Score Header} To predict more reasonably, our model take map information into consideration. Hence, cross-attention layer is used to fuse map feature into $x_{pt}$. Then we obtain a normalized score after feeding corresponding regression results into a 2-layer MLP and softmax,
\begin{center}
    $S_{pred} = SoftMax(MLP(CrossAtt_{K,M}(x_{pt}, Map))).$
\end{center}

\noindent\textbf{Temporal Flow Header} To enhance the flow of temporal information in TENET, we propose this header as an auxiliary task to realize a closed loop between history and future. Specifically, TENET regresses backward to get historical trajectories through using the prediction results, establishing temporal consistency on trajectory information. The below equation explains how we slice out future timestamps feature $x_{f}$ from $x_{pt}$ and devise a FPN (Feature Pyramid Networks) \cite{lin2017feature} module to obtain history trajectory $h_{pred}$,
\begin{center}
    $h_{pred} = {MLP}(FPN(x_f)).$
\end{center}

\begin{table*}[t]
  \caption{Results on the test set of Argoverse 2 Motion Forecasting Dataset. Brier-minFDE is the official ranking metric.}
  \label{leaderboard}
  \centering
  \begin{tabular}{llllll}
    \toprule
    Method \& Rank & \multicolumn{1}{c}{minADE ($K$=6)} & \multicolumn{1}{c}{minFDE ($K$=6)} &  \multicolumn{1}{c}{Miss Rate ($K$=6)} & \multicolumn{1}{c}{brier-minADE ($K$=6)} &  \multicolumn{1}{c}{brier-minFDE ($K$=6)} \\
    \midrule
    TENET $1^{st}$ (Ours) & \multicolumn{1}{c}{\textbf{0.70}} & \multicolumn{1}{c}{1.38} &  \multicolumn{1}{c}{0.19} & \multicolumn{1}{c}{\textbf{2.15}} &  \multicolumn{1}{c}{\textbf{1.90}} \\
    OPPred $2^{nd}$ & \multicolumn{1}{c}{0.71} & \multicolumn{1}{c}{\textbf{1.36}} &  \multicolumn{1}{c}{0.19} & \multicolumn{1}{c}{2.18} &  \multicolumn{1}{c}{1.92} \\
    GANet $3^{rd}$ & \multicolumn{1}{c}{0.73} & \multicolumn{1}{c}{\textbf{1.36}} &  \multicolumn{1}{c}{\textbf{0.17}} & \multicolumn{1}{c}{2.26} &  \multicolumn{1}{c}{1.98} \\
    Polkach $4^{th}$ & \multicolumn{1}{c}{0.71} & \multicolumn{1}{c}{1.39} &  \multicolumn{1}{c}{0.19} & \multicolumn{1}{c}{2.30} &  \multicolumn{1}{c}{2.00} \\
    QCNet $5^{th}$ & \multicolumn{1}{c}{0.76} & \multicolumn{1}{c}{1.58} &  \multicolumn{1}{c}{0.24} & \multicolumn{1}{c}{2.41} &  \multicolumn{1}{c}{2.14} \\

    \bottomrule
  \end{tabular}
\end{table*}

\subsection{Loss function}
The loss function of the model consists of three parts.
\begin{center}
    $L\text{=}{{L}_{reg}}+{{\beta}_{1}}{{L}_{score}}+{{\beta}_{2}}{{L}_{tf}},$
\end{center}
where ${{\beta}_{1}}=0.3,{{\beta}_{2}}=0.3$.
    
For ${{L}_{reg}}$, in order to construct the connection between trajectories and scores, we use GMM (Gaussian Mixture Model) \cite{reynolds2015gaussian} loss which ensures each trajectory has a reasonable score. We use all attributes of trajectory as ground truth label, including position, orientation, and instant velocity. In addition, both future and history timestamps are supervised, aiming to learn a motion pattern from history trajectory through Autoencoder \cite{liou2014autoencoder}.
\begin{center}
    ${{L}_{reg}}=-\log \sum\limits_{k=1}^{K}{{{e}^{\log ({{s}_{i}})-\frac{1}{2}\sum\limits_{t=1}^{T}{{{(A_{t}^{gt}-{{A}_{i,t}})}^{2}}}}}},$
\end{center}
where ${{A}_{i,t}}=[{{x}_{i,t}},{{y}_{i,t}},{{\cos }_{i,t}},{{\sin }_{i,t}},{{v}_{i,t}}]$, ${{s}_{i}}$ is the score of $i^{th}$ predicted trajectory.
    
For  ${L_{score}}$, to make the positive trajectory (defined as the closest trajectory to ground truth) more confident, we adopt the max-margin loss:
\begin{center}
    ${{L}_{score}}=\frac{1}{K-1}\sum\limits_{i=1,i\ne \overline{i}}^{K}{\max (0,{{s}_{i}}+}\sigma -{\Bar{{s}_{{i}}}}),$
\end{center}
Where $\sigma$ is the margin and we set it 0.15 in our loss function and $\Bar{{s}_{{i}}}$ is the score of positive prediction.

For ${{L}_{tf}}$, we take the historical trajectory ${h}_{gt}$ as the ground truth for \textbf{Temporal Flow Header} middle-level supervision.
\begin{center}
        ${{L}_{tf}}=MSELoss({{h}_{pred}},{{h}_{gt}})$
\end{center}

\begin{figure*}
	\centering
  \subfloat{\includegraphics[width = 0.33\textwidth]{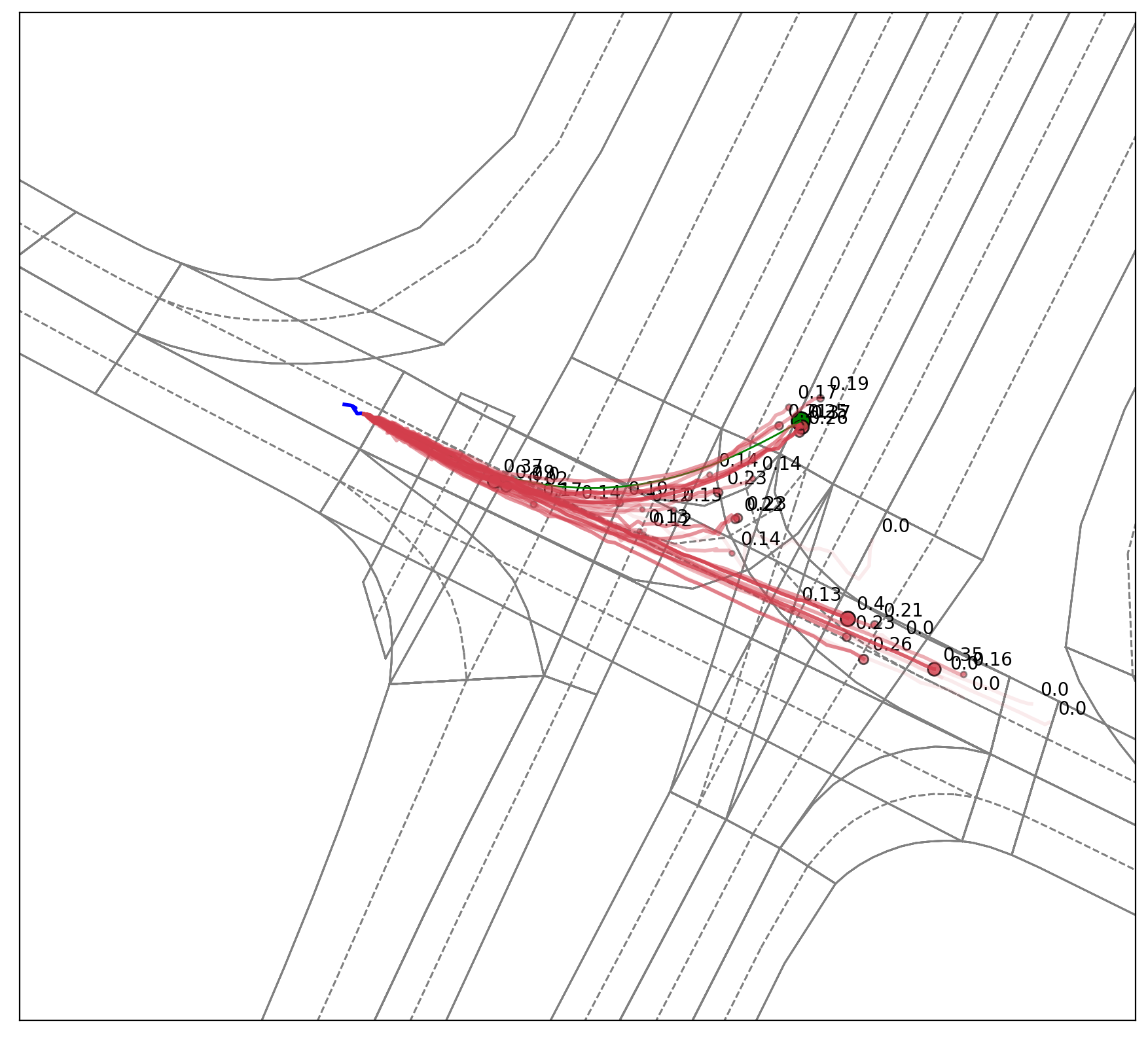}}
  \subfloat{\includegraphics[width = 0.33\textwidth]{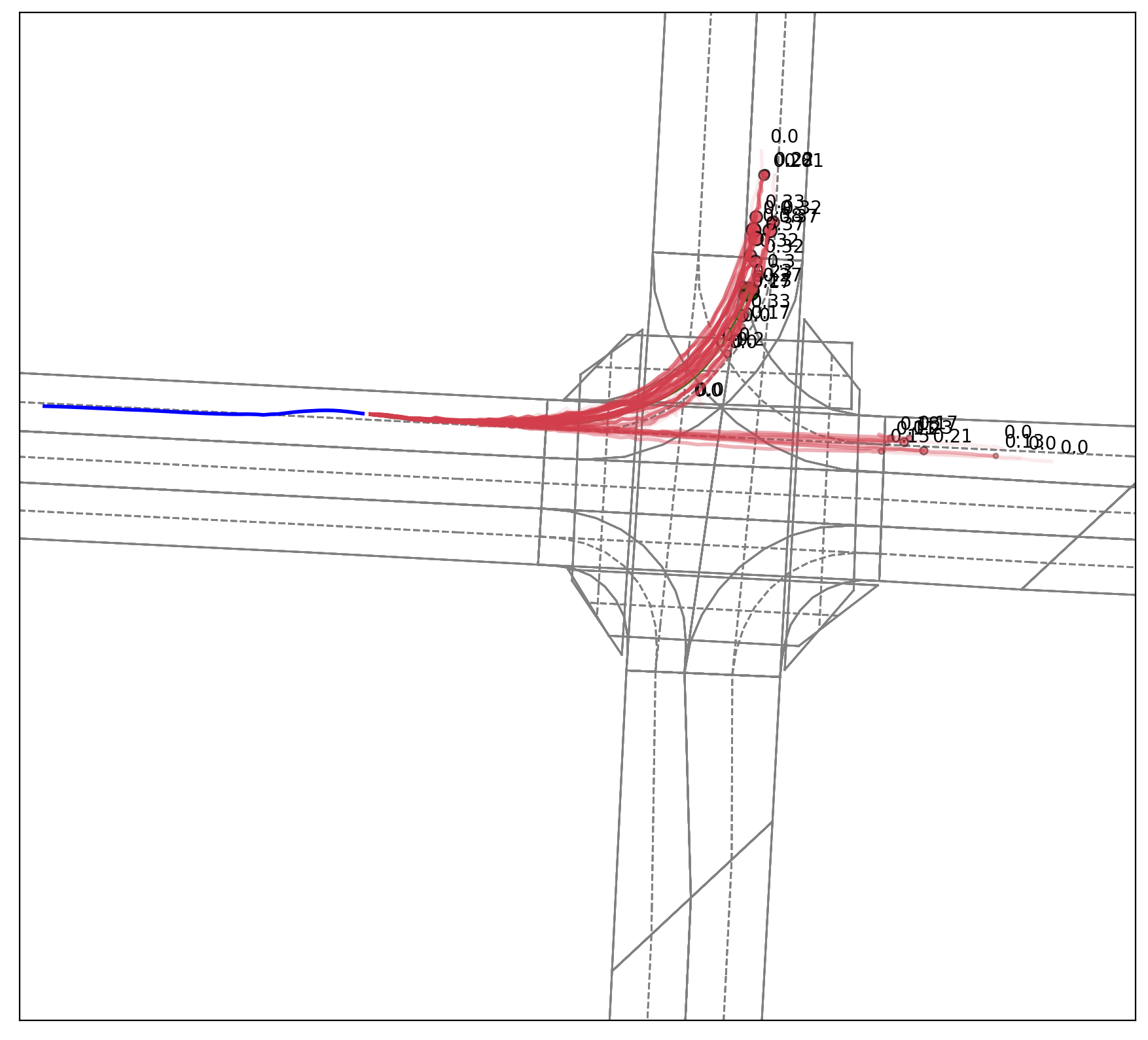}}
  \subfloat{\includegraphics[width = 0.33\textwidth]{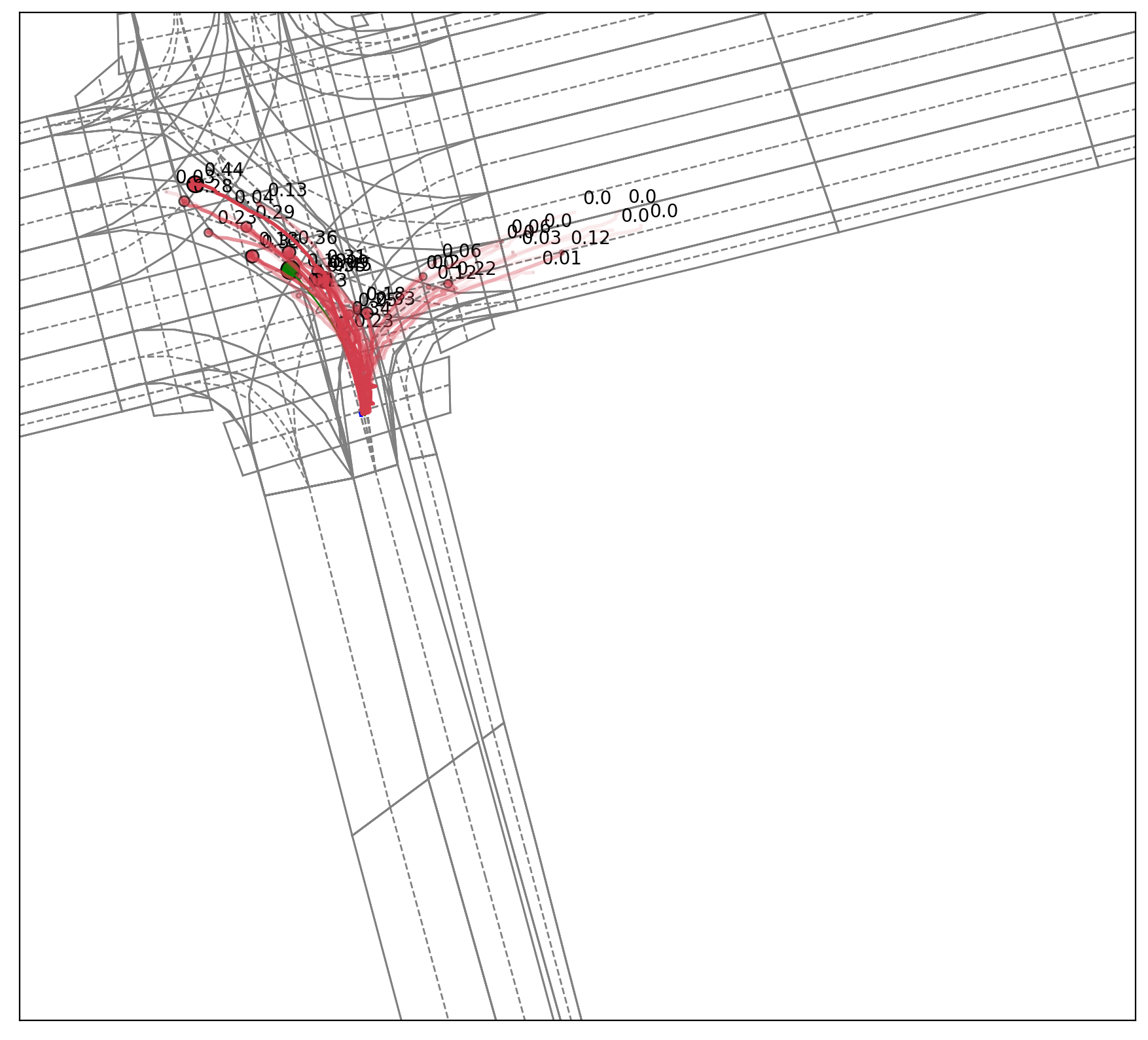}}
  \quad
  \subfloat{\includegraphics[width = 0.33\textwidth]{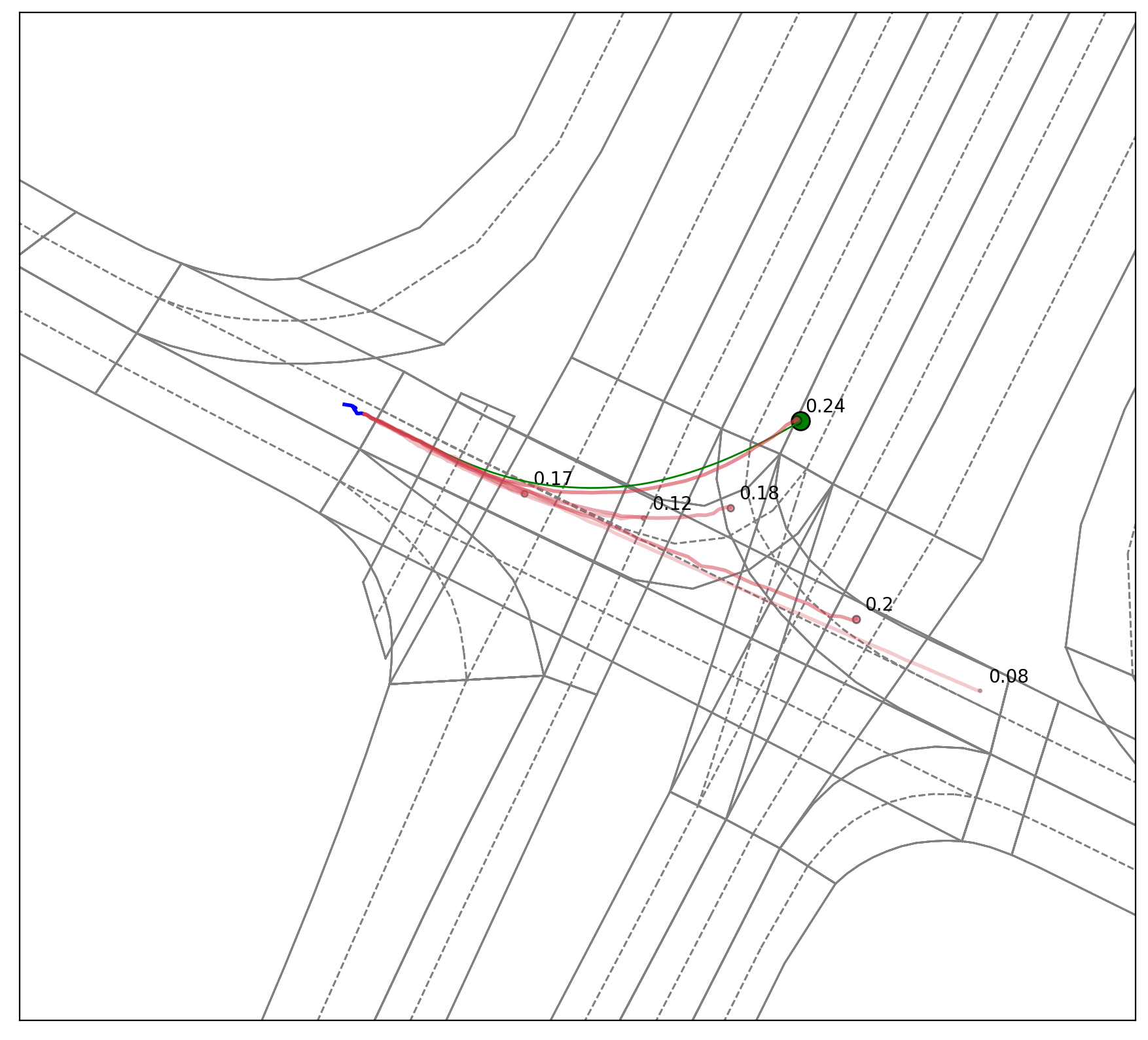}}
  \subfloat{\includegraphics[width = 0.33\textwidth]{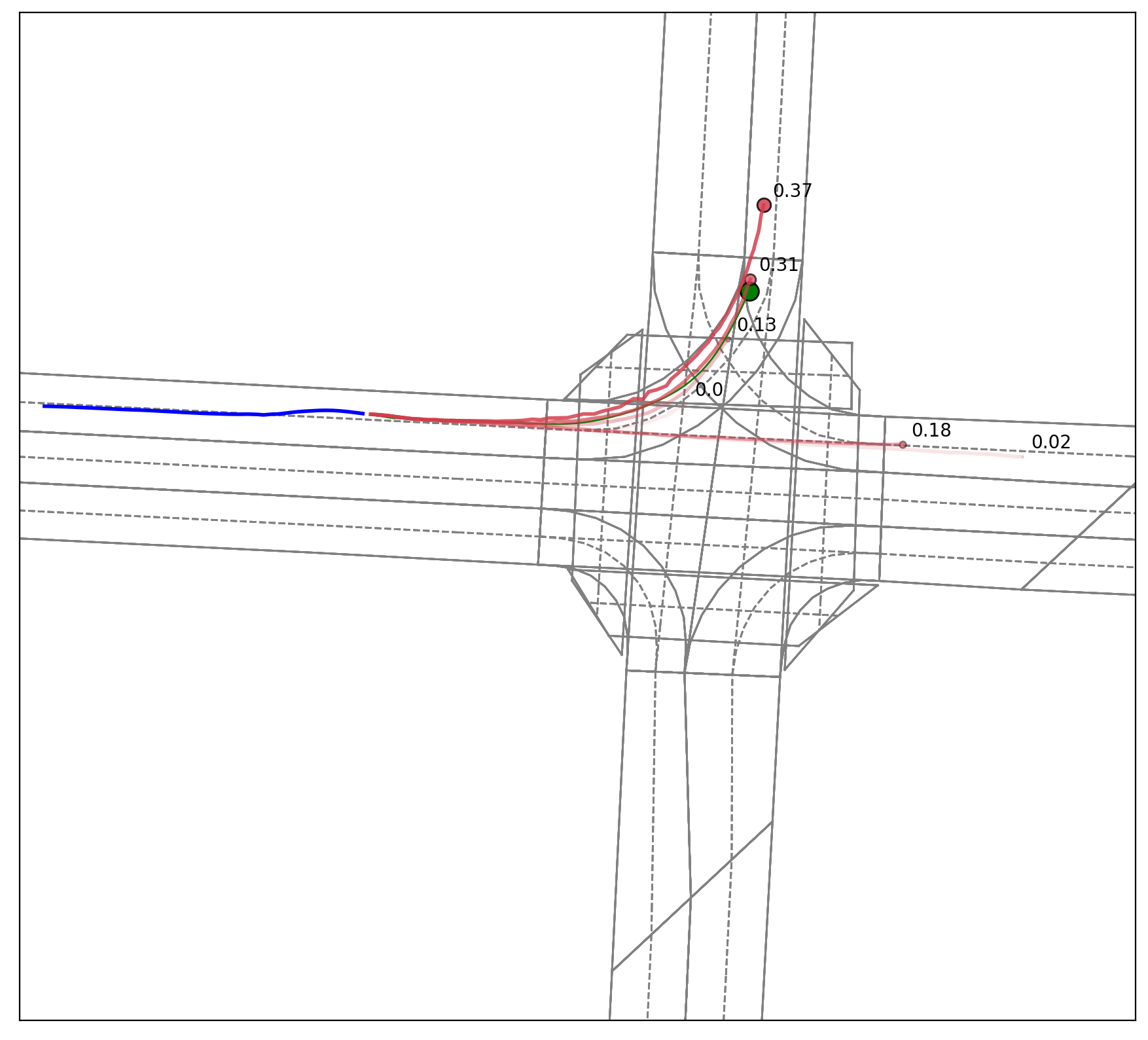}}
  \subfloat{\includegraphics[width = 0.33\textwidth]{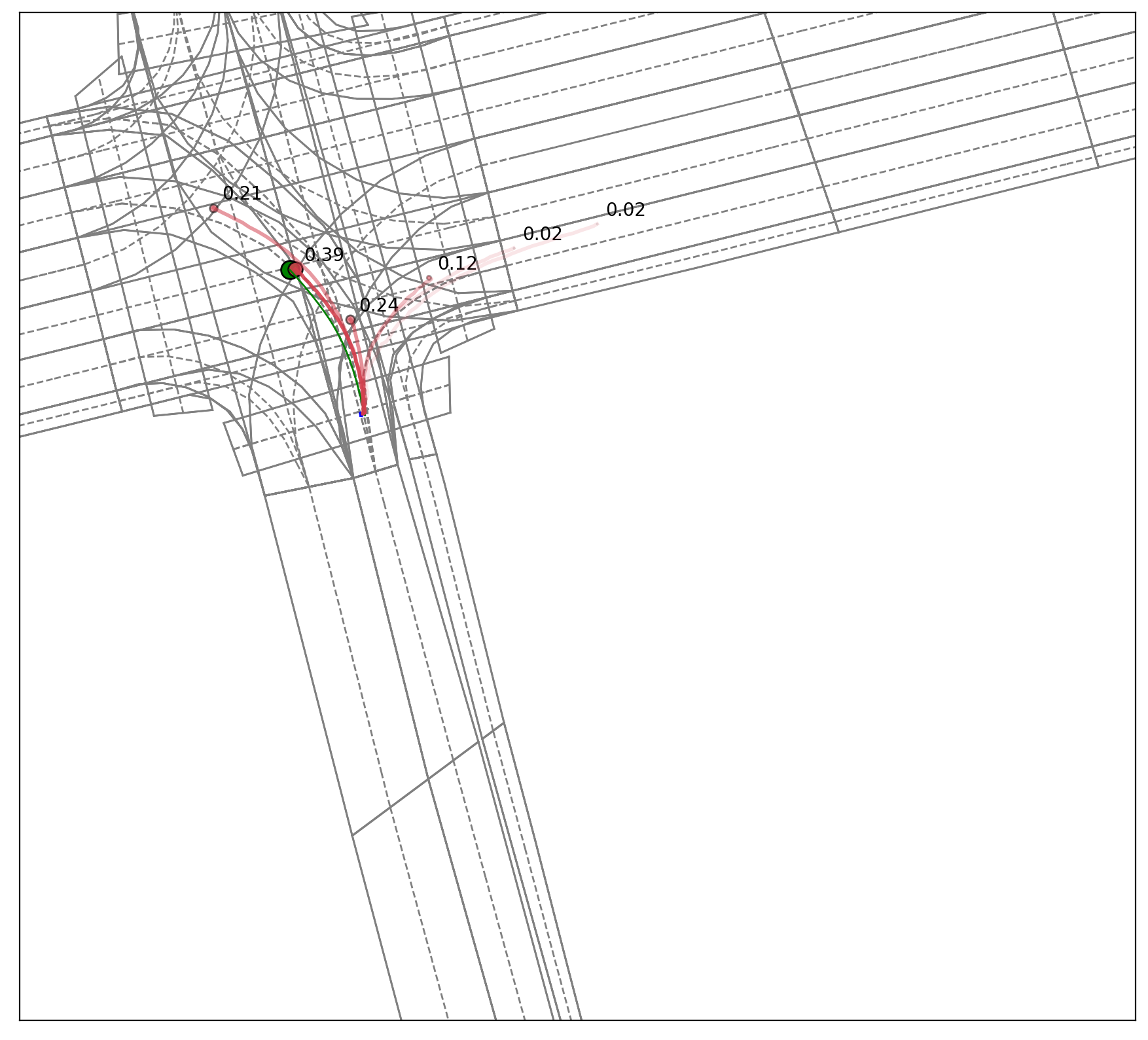}}
	\caption{\textbf{Multi-Trajectory Ensemble Visualization.} The top row shows the trajectories before ensemble and the bottom row shows the trajectories after ensemble. History trajectories are shown in blue, predicted trajectories are shown in red, ground truth trajectories are shown in green.}
	\label{ensemble}
\end{figure*}

\subsection {Data Augmentation}
	\ Augmentation, which is essential for our model, can be divided into agent augmentation and training augmentation.
	
	\ Firstly, we regard other agents as the target agent in agent augmentation. Owing to our transformer-based information interactor, the target agent and the other agents are the same during the training process. Therefore, in order to generate more kinds of conditions under one single scenario, we propose agent augmentation exchanging the identities between target agent and other agents for prediction model.
	
	\ Besides, during training, four augmentation methods (translation, rotation, flipping, and resizing) are used to generate abundant scenarios. Augmentations are applied simultaneously in agent tensors, ground-truth tensors and road graph tensors. In translation augmentation, we first generate random distance within $-3m$ to $3m$, and then translate the coordinates in the above three tensors, such as the coordinates of trajectories and centerlines etc. In rotation augmentation, we randomly rotate the coordinate system within $ [-\frac{\pi}{6},  \frac{\pi}{6}]$. In flip augmentation, we flip the coordinate system along the $y$ axis with a certain probability during training process. In resize augmentation, the entire scene are scaled by a randomly selected constant within $[0.8, 1.2]$. 

\subsection{Hard Mining}
    \ We use hard mining technique to improve the prediction of the model in difficult scenarios. Specifically, we train a proxy model with a randomly sampled training subset from the original training set, and let this proxy model perform inference on the remaining training set. Then, we mine those scenarios in which the proxy model performs poorly (scenarios with a large brier-minFDE) and increase the proportion of these scenarios in the training phase.
    
\subsection{Multi-Trajectory Ensemble}
    \ Multi-modality is a central characteristic of the trajectory prediction task. Most methods avoid unimodal prediction output using the winner-takes-all (WTA) \cite{lee2016stochastic}, which is unstable due to network initialization. Inspired by DCMS \cite{ye2022dcms}, we enhance the multi-modality of predicted trajectories by Multi-Trajectory Ensemble.
    
    \ Specifically, we use models with different random seed initializations, different degrees of hard mining, and different training epochs to generate $M$ sets of trajectories (each set contains $K$ trajectories). Then, from the total $M * K$ trajectories, we apply K-means algorithm \cite{macqueen1967some} to generate $K$ trajectory clusters. For each cluster, we average all trajectories in the cluster to generate the output trajectory, and use the sum of their scores as the score of that output trajectory. Finally, all $K$ scores of trajectories will be linearly normalized. It turns out that summing over scores is better than averaging over scores, as summing tends to give higher scores to clusters with more trajectories. We use the endpoint distance as the distance metric when clustering. Fig. \ref{ensemble} shows the visualization of Multi-Trajectory Ensemble. We also find out that the final result can be used as teacher during knowledge distillation to make single models achieve better precision. 
    
\section{Experiments}
\subsection{Dataset and Metrics}
\noindent\textbf{Dataset} The Argoverse 2 Motion Forecasting Dataset \cite{wilson2021argoverse} contains 250000 11-second scenarios with a sampling rate of 10HZ. For the training and the validation sets, the first five seconds of each scenarios are used as input and the other six seconds are used as the ground truth for models to predict. For the test set, only the first five seconds are provided. Argoverse 2 Motion Forecasting Dataset provides rich map information and contains five different dynamic categories. 

\noindent\textbf{Metrics} Argoverse 2 Motion Forecasting Challenge chooses brier-minFDE ($K$=6) as the metrics. MinFDE($K$) is the minimum displacement between $K$ final positions and the ground truth final position. Similarly, brier-minFDE multiplies ${(1.0-p)^2}$ with the endpoint L2 distance, where $p$ corresponds to the probability of the best forecasted trajectory.

\subsection{Implementation Details}
We train our model for 200 epochs (around 96 hours) using eight 2080Ti GPUs. As for the input, we sample actors and lanes with distance less than 100 meters from the target agent. Using rotation and translation, each scene is normalized using the target agent as the center. Specifically, the most recent historical position of the target agent (at the $49^{th}$ frame) is taken as the origin, and the direction of the target agent historical trajectory is aligned with the positive axis of $X$. We use Adam \cite{kingma2014adam} optimizer with an initial learning rate of 2.5e-4, which is decayed to 2.5e-5 at 170 epochs and to 2.5e-6 at 190 epochs. Agent dimension (denoted by $A$) is set to 32, and map dimension (denoted by $M$) is set to 128 at training time. Since the model supports dynamic agent and map dimensional inputs, the dimensions increased to 64 and 256 respectively during testing. We use padding and clipping to align the dimensions. All transformer modules in TENET contain 128 hidden units.

\subsection{Results}
    \ \textbf{Argoverse 2 Motion Forecasting Competition.} We evaluate TENET on Argoverse 2 Motion Forecasting Competition. As shown in Table \ref{leaderboard}, our method ranks $1^{st}$ on the final leaderboard. The official metric is brier-minFDE. 
    
    \ To demonstrate the effectiveness of Multi-Trajectory Ensemble, we compare the average performance of trajectories before ensemble and after ensemble on the Argoverse 2 Motion Forecasting test set. As shown in Table \ref{ablation} and Figure \ref{ensemble}, Multi-Trajectory Ensemble integrates all trajectories and enhances the multi-modality and confidence of the prediction.  
    
    \ Besides, Table \ref{ablation} shows that increasing the input size yields better prediction results. So we reduce the input size in the training phase and increase the size in the testing phase to accelerate training while maintaining predictive accuracy.
    
\begin{table}
  \caption{Ablation study results of modules on the Argoverse 2 test set.}
  \label{ablation}
  \centering
  \begin{tabular}{llll}
    \toprule
    Method &  \multicolumn{1}{c}{big input size} &  \multicolumn{1}{c}{MTE} &  \multicolumn{1}{c}{brier-minFDE ($K$=6)} \\
    \midrule
    \multirow{3}{*}{TENET} & \multicolumn{1}{c}{ } & \multicolumn{1}{c}{ } & \multicolumn{1}{c}{2.03}  \\
      & \multicolumn{1}{c}{$\surd$} & \multicolumn{1}{c}{ } & \multicolumn{1}{c}{2.01}  \\
      & \multicolumn{1}{c}{$\surd$} & \multicolumn{1}{c}{$\surd$} & \multicolumn{1}{c}{\textbf{1.90}}  \\

    \bottomrule
  \end{tabular}
\end{table}

\section{Conclusions}
This technical report presents an effective method for the motion prediction task. We develop an efficient Transformer-based network to predict trajectories, and we propose \textbf{Temporal Flow Header} to enhance the trajectories. Besides, we devise a training strategy to accelerate model training and a strong K-means based ensemble method. We conduct experiments on Argoverse 2 Motion Forecasting Dataset \cite{wilson2021argoverse} and achieve \textbf{\textit{state-of-the-art}} performance. Finally, we hope this work will be a strong baseline in this motion prediction task.
{\small
\bibliographystyle{ieee_fullname}
\bibliography{egbib}
}

\end{document}